Title:

# Machine vision detection to daily facial fatigue with a nonlocal 3D attention network

Authors:

Zeyu Chen,[1,2] Xinhang Zhang,[1,2] Juan Li[1,2], Jingxuan Ni[1,3], Gang Chen[4], Shaohua Wang[4], Fangfang Fan[5], Changfeng Charles Wang[6*], Xiaotao Li,[1,2,7,8 # *]

1. BIAI Intelligence Biotech Co., Ltd, Shenzhen 518055, GD, China; 2. BIAI Inc., Chelmsford, MA 01824, USA. 3. Department of Mathematics, School of Science, Shantou University, Shantou, GD, China. 4. Shenzhen Breo Technology CO., LTD., Shenzhen, GD, China. 5. Department of Neurology, Harvard Medical School, Harvard University, Boston MA, USA. 6. Stealth Mode Startup, INC., Chicago, Illinois, USA. 7. Shenzhen Key Lab of Neuropsychiatric Modulation, Guangdong Provincial Key Laboratory of Brain Connectome and Behavior, CAS Key Laboratory of Brain Connectome and Manipulation, CAS Center for Excellence in Brain Science and Intelligence Technology, the Brain Cognition and Brain Disease Institute, Shenzhen Institutes of Advanced Technology, Chinese Academy of Sciences; Shenzhen-Hong Kong Institute of Brain Science-Shenzhen Fundamental Research Institutions, Shenzhen, 518055, China. 8. Department of Brain and Cognitive Sciences, Massachusetts Institute of Technology, Cambridge, MA, USA.

*Corresponding author e-mail: xtli@mit.edu (X. Li)

**Abstract:**

Fatigue detection is valued for people to keep mental health and prevent safety accidents. However, detecting facial fatigue, especially mild fatigue in the real world via machine vision is still a challenging issue due to lack of non-lab dataset and well-defined algorithms. In order to improve the detection capability on facial fatigue that can be used widely in daily life, this paper provided an audiovisual dataset named DLFD (daily-life fatigue dataset) which reflected people's facial fatigue state in the wild. A framework using 3D-ResNet along with non-local attention mechanism was training for extraction of local and long-range features in spatial and temporal dimensions. Then, a compacted loss function combining mean squared error and cross-entropy was designed to predict both continuous and categorical fatigue degrees. Our proposed framework has reached an average accuracy of 90.8% on validation set and 72.5% on test set for binary classification, standing a good position compared to other state-of-the-art methods. The analysis of feature map visualization revealed that our framework captured facial dynamics and attempted to build a connection with fatigue state. Our experimental results in multiple metrics proved that our framework captured some typical, micro and dynamic facial features along spatiotemporal dimensions, contributing to the mild fatigue detection in the wild.

**Keywords**: fatigue detection, in the wild, DLFD dataset, non-local attention, 3D-ResNet

# 1 Introduction

Fatigue is one of psychophysiological conditions with a feeling of physical or mental tiredness, usually caused by exertion and then characterized by a decreased capacity for physical activity and mental processing [1]. It can be roughly categorized to physical and mental parts, however, physical fatigue mainly involves the over-taxation of the muscular system while mental fatigue is a transient decrease in cognitive performance mainly induced by excessive demands on cognitive systems [2]. Mental fatigue, also known as cognitive fatigue, often leads to a lowered productivity and an increased risk of safety [3]; and long term of mental fatigue (2-3 weeks) can even be as one of major symptoms for major depression [4]. Hence, in-time detection of mental fatigue will largely contribute to the balance of work arrange and life satisfaction and even prevent fatigue-related brain disorders.

In terms of facial fatigue detection, people can perceive their own fatigue and the fatigue can be also observed by others through some typical features on their faces such as droopy eyes, long blinks and yawning [5]. PERCLOS (percentage of eyelid closure over the pupil over time) [6] and eye aspect ratio [7] are both widely used as metrics to judge drowsiness level via measuring the degree of eye closure and the frequency of eye blink. And statistical analysis can be applied based on these drowsiness-related indicators to measure individuals' fatigue level, which have been widely applied in some situations like driving monitor system (DMS) [8]. However, those traditionally statistical methods including haar-like feature [9] and Adaboost classifier [10] are limited to detect a severe state of fatigue. In terms of detection for mild and modern fatigue, deep learning algorithms especially convolutional neural networks (CNN) have already showed some advantages over traditional ones [11]. To develop a better detection for facial fatigue in daily life, two major trends with computer vision have showed in the current facial fatigue detection: one is a transition from image-based detection [12-16] to video-based detection [17-22], and the other is from experimentally simulated data to the data collected in the real-world scenario. Though the existing methods have achieved up to 97% accuracy [18], most of them used well-known fatigue datasets like NTHU-DDD [23] and YawDD [24], which were either simulated in the laboratory or collected at vehicle-mimicking conditions [25]. Those methods were sufficiently effective when there were distinct features between fatigue and alert state. However, in many situations in real life the fatigue is not obvious so that facial fatigue detection especially for mild fatigue is still a challenge in the wild.

In order to improve fatigue detection level that it can be widely used in different life scenarios in the wild, we proposed an audiovisual dataset named DLFD (daily-life fatigue dataset), containing people's natural facial status at different conditions. Furthermore, we employed paradigms of 3D CNN and non-local attention blocks to increase the ability of our models in recognizing mild fatigue in the wild. Through the experiments on dataset of DLFD, we proved the effectiveness and advantages of our model.

## 2 Materials and Methods

### 2.1 Dataset Properties and Description

DLFD dataset consists of 375 videos with about 20,000 frames length of which varies from 5 seconds to 2 minutes. The mean resolution for these videos is approximately 1056 x 629 and mean frame rate is 30 fps. The facial videos in the dataset are self-captured by 32 volunteers when they are in different life scenarios such as awaking, working, off work, breaking, etc. and at different time from morning to midnight. To some extent, this collection recorded people's natural rather than faked fatigue state as much as possible, thus accurately reflected people's fatigue degree in the wild. DLFD dataset was labeled independently by three psychology-based annotators based on subjectives' self-evaluation. According to the visual and acoustic information contained in volunteers' video, annotators evaluated their degree of fatigue ranging from 1 to 5. 1 represents very significant alert state and 5 represents very significant fatigue state. We calculated two types of labels, one of which for continuous task and the other for categorical task. The former one is calculated by the mean evaluation of three annotators and the latter one is deduced from the mean evaluation by isometric binning techniques. For simplicity, we call them continuous and categorical label respectively.

## 2.2 Network Architecture

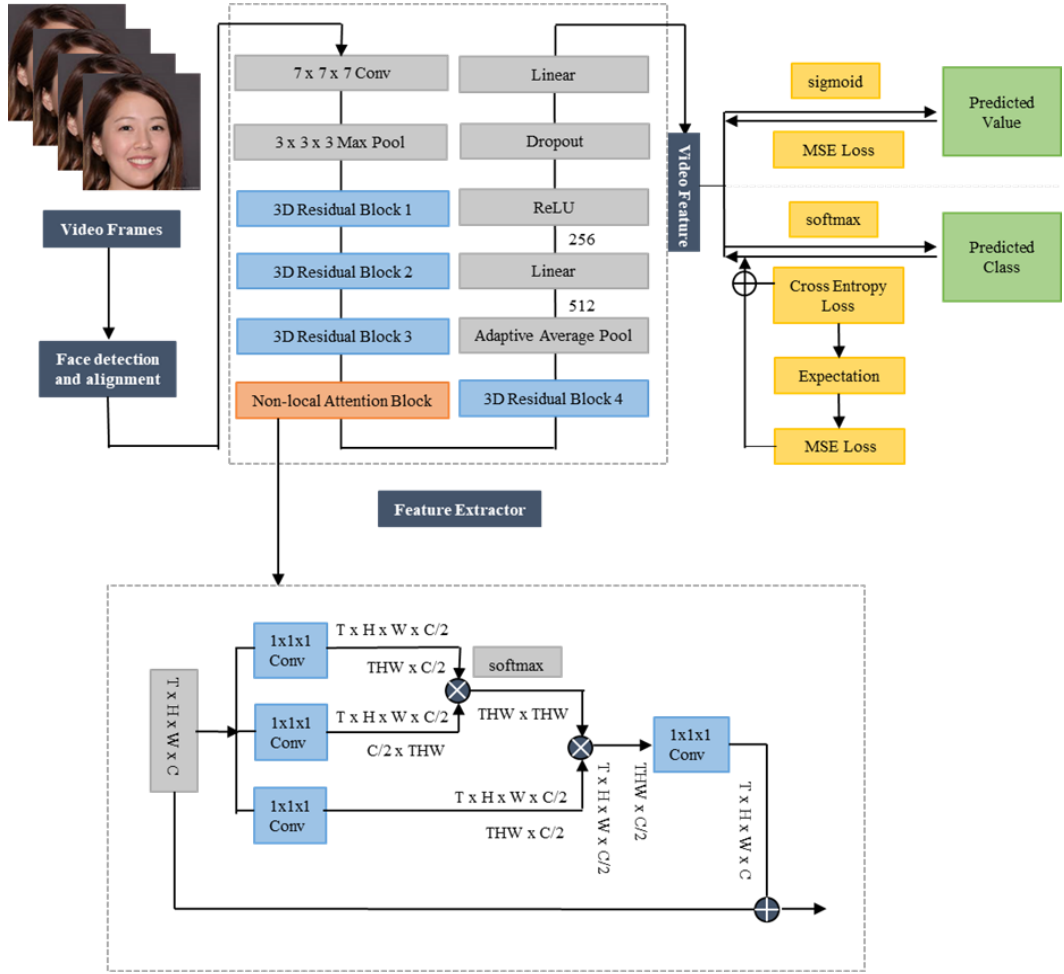

Figure 1. **The architecture of our proposed framework for fatigue detection on video frames.** (a) Videos are firstly divided into frames, cut out the face region and aligned. Then they are fed into modified ResNet-18 which consists of 3D convolutional blocks, 3D pooling blocks, 3D residual blocks, 3D non-local attention block and linear blocks to extract feature. Two methods are designed to predict the fatigue value and calculate loss value for backpropagation. One uses MSE loss to predict continuous value and the other uses MSE + CE loss to predict categorical value. "⊕" denotes summation for MSE and CE loss. (b) The non-local attention block is detailed in the dotted box. T, H, W, C represent time (number of frames), height, width and channel. "⊕" denotes element-wise sum and "⊗" denotes matrix multiplication. This block is placed below the third 3D residual block. Thus, the dimension for the input of non-local block is 4 x 14 x 14 x 256.

We employed the paradigm of ResNet-18 as the backbone model and turned the 2D model into 3D convolution network based on the convolutional kernel inflation technique in [26]. In addition, taking advantage of non-local attention mechanism, we incorporated it below the third basic block in 3D-ResNet. Noted that different from the single linear layer design in classic ResNet, we appended two fully connected layers after convolution blocks. It is noteworthy that one of our novel works here was the adoption of both mean squared error loss function and cross entropy loss function in fatigue level prediction. This strategy was motivated by construction of loss function in multi-task learning and realized by an expectation transformation technique, which was illustrated below in the section of loss function. The softmax layer was first used to categorize the features extracted by linear blocks and through expectation transformation, the categorical prediction could be transformed to a continuous value for the calculation of MSE loss. The overall network architecture is presented by Figure 1.

## 2.3 Non-local Attention Module

The non-local attention has been viewed as a way to take global information into consideration in constructing feature map, showing its advantage in many applications of computer vison like image restoration and people reidentification [27, 28]. According to the work of Wang et al. [26], here we give a general introduction to non-local attention block. Following their work, the non-local operation on convolutional networks can be defined by:

$$y_i = \frac{1}{C(x)} \sum_{\forall j} f(x_i, x_j) g(x_j) \qquad (2.4.1)$$

The equation above shows the calculation of output features $y_i$ given input features $x_i$. Here $i$ represents the index of current response (i.e., feature) from either spatial or temporal dimension and $j$ represents the index of global response from all dimensions. Function $f$ measures the affinity between the i-th feature and the j-th feature, and function $g$ projects the given features $x$ into a feature space. $C(x)$ is a normalization factor. In our work, we adopted the embedded gaussian form mentioned in [26] in calculating $f$ since it is a natural extension of the self-attention blocks widely used in sequence models. Here is the architecture of our non-local attention module:

The non-local attention module tackles long-term dependencies in spatial and temporal

dimensions from four major steps:

- 1 x 1 convolution is adopted to respectively generate two feature spaces $u(x), v(x)$ given input feature $x$, which are defined as: $u(x) = W_u x, v(x) = W_v x$. The autocorrelation of features is calculated by dot product of feature spaces i.e., $u(x)^T v(x)$. Here $u(x_i)^T v(x_j)$ represents the relationship between i-th feature and j-th feature.

- Considering the embedded gaussian format of $f$ in equation (2.4.1) and the need of normalization, outputs in step 1 pass softmax to form attention map. The attention weight from the i-th feature to the j-th feature is given by:

$$\alpha_{ij} = \frac{\exp(s_{ij})}{\sum_{j=1}^{N} \exp(s_{ij})}, s_{ij} = u(x_i)^T v(x_j) \qquad (2.4.2)$$

- Based on the attention weight, original feature $x$ is first projected to another feature space through function $g$, and then summed up along j-th dimension to produce self-attention feature map. Given activation function A, the self-attention feature $y_i$ is calculated by:

$$y_i = A(\sum_{\forall j} \alpha_{ij} g(x_j)) \qquad (2.4.3)$$

- Motivated by the method in residual network, the self-attention feature and original feature are synthesized to calculate the non-local output $z_i$:

$$z_i = \gamma_z y_i + x_i \qquad (2.4.4)$$

Noted that $\gamma_z$ is trainable parameter which is initialized by zero. It indicates that in the beginning of training, local information generated by the convolution is more heavily relied, and gradually non-local information is introduced to grasp information of larger scope.

## 2.4 Loss Function

In order to gain continuous value for fatigue measurement, at the first stage of our research we used mean square error as loss function and added a sigmoid layer in the end to achieve predictions between 0 and 1. However, it should be pointed out that using only mean squared error as loss function has some defects. Due to the mathematical property of the derivative derived by combining sigmoid with MSE, optimization might be easy to fall into a local minimum. Additionally, we also preferred to train a categorical model so that it could be compared with other state-of-the-art models. During the design of this categorical model, since both categorical and continuous predictions for evaluating fatigue degree were expected to obtain through the network, we designed a loss function which combines MSE and cross entropy loss together by the inspersion of multi-class learning. The motivation behind this idea is instinct that the featured extracted by networks can be activated by sigmoid function and evaluated by cross entropy loss to avoid gradient problems caused by the combination of sigmoid and MSE. Also, since the information is more detailed when using continuous label, the introduction of MSE loss helps the network better learn the matching rules between features and labels, thereby reducing the difficulty of learning. The challenge in this part is the continuous prediction for fatigue given the output of sigmoid layer. Here we borrowed the method from the work of Ruiz et al. [29] and designed a strategy named expectation transformation which transformed the probability of fatigue from sigmoid layer to a continuous value between 0 and 1 representing fatigue prediction. Specifically, the fatigue value from 0 to 1 can be binned to two intervals, [0, 0.5) and [0.5, 1]. The probability that the fatigue value falls within the corresponding interval is given by the sigmoid output and the midpoints of these intervals can be viewed as their representation. In this way, mathematical expectation defined in discrete random variable can be calculated and regarded as the continuous prediction.

Noted that our proposed expectation transformation strategy was easily generalized to multi-class task, so the usage of our model architecture is not limited to binary-class situation. Here we derive the mathematical form of our expectation transformation in multi-class situation. Supposed that we have $k$ neurons in softmax layer which will predict probabilities for $k$ classes. Given sample $j$ and class $i$, $q_{ji}$ is the predicted probability by softmax layer. The fatigue value will be divided into different bins determined by $k:[\frac{i}{k},\frac{i+1}{k})$, $i=1,...,k-1$.

Using the midpoints of these bins, the continuous prediction of fatigue can be given by:

$$E(q_j) = \sum_{i=1}^{k} q_{ji} \frac{2i+1}{k} \qquad (2.5.1)$$

The loss function of our model is defined by:

$$L = \alpha L_{softmax} + L_{mse} \qquad (2.5.2)$$

Here $L_{softmax}$ represents the cross-entropy loss and $L_{mse}$ represents the mean squared error loss. $\alpha$ is a hyperparameter which adjusts the weight of cross entropy loss function and further influences the learning effect of network. Supposed that the ground truth label of the category version is $p_{ji}$, which is 1 or 0 generated by one-hot encoding and the continuous ground truth label is $y_j$ ranging from 0 to 1. The concrete formula of $L_{softmax}, L_{mse}$ are given by:

$$L_{softmax} = -\frac{1}{N}\sum_{j=1}^{N}\sum_{i=1}^{k} p_{ji} \log q_{ji}, \quad L_{mse} = \frac{1}{N}\sum_{j=1}^{N}\left[y_j - E(q_j)\right]^2 \qquad (2.5.3)$$

For simplicity, in this paper we called this loss function MSE + CE loss which combined mean squared error and cross entropy loss together.

## 2.5 Implementation Details

At the beginning, we pre-processed our data by splitting videos into continuous frames and then using MTCNN [30] for face detection and alignment. Since DLDF dataset was collected by different volunteers in multiple in-the-wild scenes, face detection and alignment were important for getting standard data format and stable feature positions. MTCNN can detect face with a bounding box and estimate the positions of five representative facial landmarks: left eye center, right eye center, nose center, left corner of mouth and right corner of mouth. Given the above information, the affine transformation was applied to align the five landmarks to the same position roughly with rotation and scaling. The final frame of each face ensured the similar landmark positions, face size and the same image size 224 x 224. The input for our network is a 4-dimensional tensor and each tensor represents a small clip from each video. Considered the computing capability, every 32 frames were divided from each video and assigned to each tensor.

Further, we implemented our framework by PyTorch and NVIDIA GeForce RTX 3070 and

GeForce RTX 2080Ti. We transferred a 2D ResNet-18 model pretrained on MS-Celeb-1M face recognition dataset [31] to 3D model by inflating the kernels [32]. Specifically, each of the t dimension of 3D kernel t x k x k are initialed by pre-trained k x k weights of the corresponding 2D kernels. For our framework with MSE loss, we used the total 375 samples in DLFD dataset, and trained with a starting learning rate of 0.00001, weight decay of 0.1, batch size of 32 and use Adam with default values as the optimizer. We used learning rate decay and given a certain number of iterations, the learning rate was reduced by 0.2 times at 1/3 and 2/3 respectively. Additionally, we resized the input image to 112 x 112 to make the computation complete on one GPU. Thus, the size of the final input tensor was 32 x 3 x 112 x 112. For our framework with MSE + CE loss which produced categorical prediction, we selected 69 samples and trained with the same learning rate and decay method, weight decay of 0.001, batch size of 16, Adam optimizer as well as the resized input image of 224 x 224. In this case, the size of the final input tensor was 32 x 3 x 224 x 224 (Fig 1).

For the sake of improving the generalization ability of the model and avoiding obvious overfitting, we adopted up to 6 kinds of data augmentations including random horizontal flipping, gaussian blur as well as random changes on image lightness, contrast and saturation. We composed these changes together and introduced a probability of 0.5 so that the occurrence of each augmentation can be randomized. Noticed that above augmentations were fixed and remained the same in one tensor, which was one 32-frame clip from a video, while randomized and varied among different clips. Additionally, early stop was used and twenty epochs was set as the patience.

## 3 Results

### 3.1 Comparison

During our experiments, we attempted to test multiple hyperparameter combinations and different backbone models. In addition, we compared the effect of non-local attention block on our model performances, with no attention block, attention block adding at different residual blocks, respectively. Two kinds of loss functions were used for evaluation: one was MSE loss and the other was MSE + CE loss. Unless otherwise specified, the following experiments all used the total 375 samples in DLFD dataset, the pre-trained framework with non-local attention mechanism placed below the third residual block, and default hyperparameter settings illustrated in the section of 2.5, implementation details. Each comparison adopts the controlled variable method to ensure that the hyperparameters and method settings related to model training other than the variable to be compared remain unchanged.

**3.1.1 The effect of batch size and data augmentation**

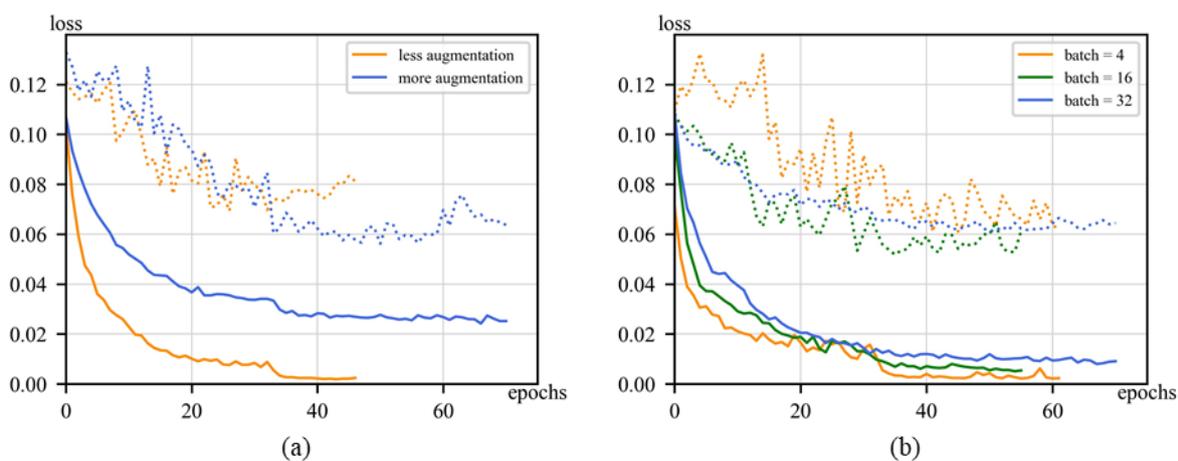

Figure 2. **Loss curves of training and validation set under different numbers of data augmentations and batch sizes using the framework with MSE loss.** The solid lines are training curves and dashed lines are validation curves. (a) Loss curves for different data augmentations. Orange line shows the change of loss values using "less augmentation", which consists of 3 augmentations including horizontal flipping, brightness increasement and brightness reduction. The augment strategy is that one of the above augmentations plus original images without augmentation is randomly selected during each batch in training. Blue line shows "more augmentation" using 6 augmentations including gaussian blur, the adjustment of color saturation and contrast and the above three. Also, probability is introduced in each transformation to control the strength. The strategy is to compose these augmentations together and introduce 0.5 probability to decide whether a transform will occur or not. With the probabilities, the richness of augmentations in "more augmentation" scenario is greatly increased. (b) Loss curves for different batch sizes.

A proper batch size and data augment strategy were significantly good for data richness and diversity (Fig 2), which were conducive to the learning of general characteristics and the generalization of model. With the framework with MSE loss, we tried batch size of 4, 16 and 32 on 100 samples, and also two data augmentation strategies with the difference on augmentation diversity (1st: 3 augmentations including horizontal flipping, brightness increasement and brightness reduction, 2nd: 6 augmentations including the above three, gaussian blur as well as the adjustment of color saturation and contrast). In each experiment, we ensured the other control variables remained the same. Figure 2 showed that the training and validation loss curve with the batch size of 32 were the smoothest and the number of epochs before early stop is the most. In addition, compared with "less augmentation" strategy of 3 augmentations, "more augmentation" strategy of 6 augmentations shortened the discrepancy of the loss between training and validation set, and prolonged iteration times before early stop as well. Hence, in the following experiments below, we all used the batch size of 32 and "more augmentation" strategy of 6 augmentations.

### 3.1.2 The effect of backbone model

Here we attempted serval network structures as our backbone from 2D models to 3D models using the framework with MSE loss. In table 1, regarding 2D-ResNet as the baseline, it reached its bottleneck on training loss between 0.065 and 0.07, with a validation loss in [0.085, 0.09]. In contrast, 3D model reduced train loss lower to 0.02 and validation loss reached around 0.06. Furthermore, 3D-ResNet with attention block and transfer learning reached the minimum loss both in training set and validation set. Additionally, the loss decreased more stable than 3D model without transfer learning, then early stop was encountered until 71 epochs.

| Network | Train loss | | Validation loss | | Total Epochs |
|---|---|---|---|---|---|
| | Max | Min | Max | Min | |
| Baseline (2D-R18) | 0.0922 | 0.0666 | 0.1028 | 0.0864 | 45 |
| 3D-R18 | 0.0710 | 0.0358 | 0.1055 | 0.0742 | 28 |
| 3D-R18 + Attention | 0.0710 | 0.0297 | 0.1310 | 0.0645 | 54 |
| 3D-R18 + Attention + Transfer | 0.1068 | **0.0242** | 0.1330 | **0.0562** | 71 |

Table 1. **Comparison of different backbone models with MSE loss.** 2D ResNet-18 is regarded as baseline model. 3D architecture, attention block and transfer learning are added successively to show the decrease of minimum loss both in training set and validation set. Columns in the table show the maximum and minimum

loss of training and validation set. Additionally, the number of total epochs before early stop is added, which shows that overfitting is not easy to trigger after adding attention block and transfer learning.

### 3.1.3 The effect of attention block

We inserted an attention block into different positions using the framework with MSE loss to study the influence of the position for attention. In this experiment, we compared the change of loss values using the framework without attention block and with attention block added after the third and fourth residual block respectively. Excepted for the change in the position of attention block, we ensured other control variables remained the same. In addition, we processed the validation loss curve with corrected exponential moving average (EMA) (Fig 3) to make the trend of loss more obvious. And we demonstrated that with the EMA validation loss curve, loss was lower when attention block was added below the fourth residual layer, compared with the condition without attention, but the overfitting began earlier at about 20-th epoch. While attention block was inserted after the third residual layer, its performance was further improved since a loss descended smoothly to a lower stage after serval epochs without encountering overfitting, though initial loss was higher.

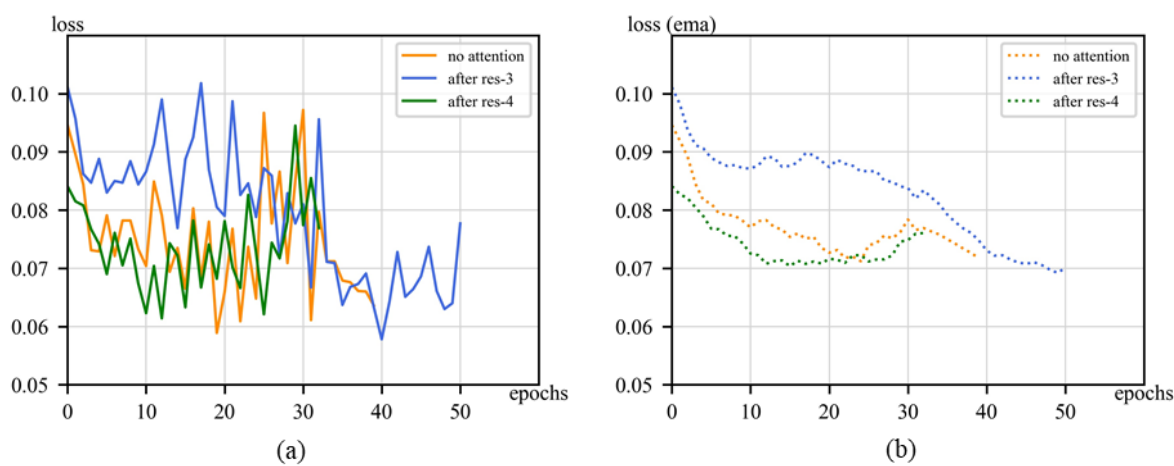

Figure 3. **Loss curves of validation set when non-local attention block is inserted into different positions in the framework with MSE loss.** That all use transfer learning with the same augmentation strategy, batch size of 32 and other default hyperparameter settings. (a) Validation loss curves for the adjustment of attention block position. Orange line shows the loss curve using the framework without attention. Blue line and green line show the network using the framework where attention block is inserted respectively after the third and fourth residual block. (b) Validation loss curves after exponential moving average for different positions of attention block. The corrected exponential moving average is used to smooth the loss curve in (a) and highlight the tendency of loss change.

| Attention Block Position | Train loss | | Validation loss | | Total Epochs |
|---|---|---|---|---|---|
| | Max | Min | Max | Min | |
| Without attention | 0.1025 | 0.0259 | 0.0946 | 0.0589 | 40 |
| After residual block 3 | 0.1117 | **0.0251** | 0.1011 | **0.0578** | 61 |
| After residual block 4 | 0.1084 | 0.0329 | 0.0808 | 0.0614 | 33 |

Table 2. **Comparison of loss values when non-local attention block is inserted into different positions in the framework with MSE loss.** That all use transfer learning with the same augmentation strategy, batch size of 32 and other default hyperparameter settings. When attention block inserts after the third residual block, the minimum training loss and validation loss both achieve the lowest.

### 3.1.4 The effect of loss function

In this part we illustrated the experimental results using the framework with MSE + CE loss. During our experiments, we found the categorical model highly sensitive to sample selection mainly due to label subjectiveness, which will be shown at the discussed part. Hence, we trained our framework on a carefully selected 69 samples and tested the model on 91 samples picked from the whole samples, based on the distinct features on fatigue levels (Fig 4). It was showed that our best performance of validation accuracy for categorical model was higher than 95%. For a deeper evaluation of the model performance, 91 samples were picked for test set. Then the fatigue group and alert group were treated as positive and negative groups in turn to calculate accuracy, precision, recall and F1 score. And the ROC curve was showed at figure 4 and the average area under curve (AUC) reached about 0.93. From table 3, the average F1 score on evaluating these samples reached at 0.84.

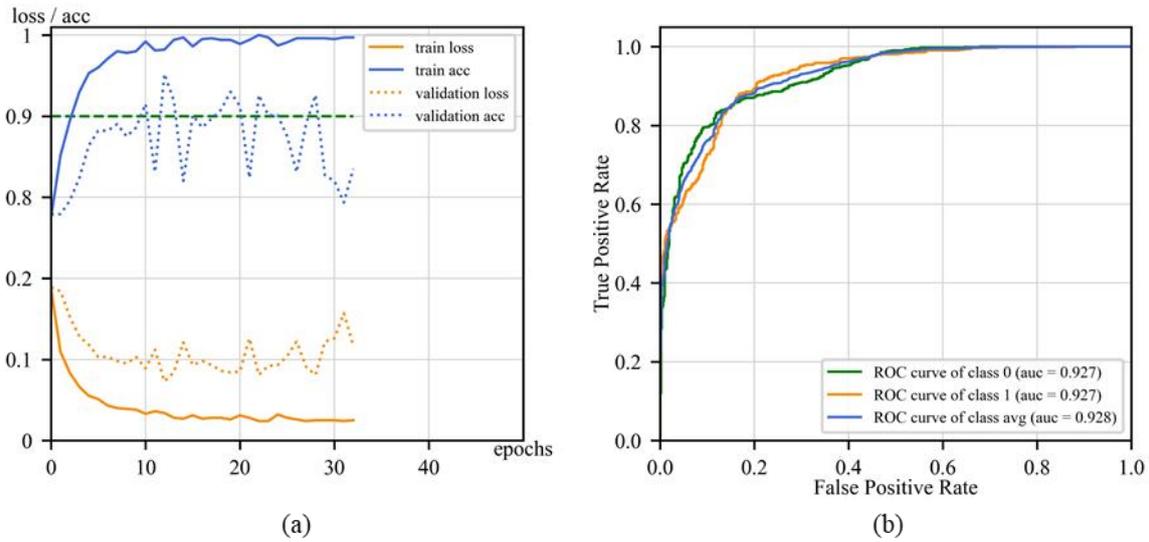

Figure 4. **Curves for the framework with MSE + CE loss.** That use transfer learning on 69 selected samples and attention block is inserted after the third residual block. (a) Loss and accuracy curves. Average accuracy on validation set surpasses 0.9, which is highlighted by a green horizontal dashed line. (b) ROC curves. That shows the ROC curves when treating fatigue label and alert label respectively as positive group. In the green curve, fatigue label is regarded as positive group and alert label is regarded as negative group, while their positions are reversed in the orange curve. Also, the average values of true positive rate and false positive rate for the above two conditions are plotted as the blue curve. The values of AUC in the above three conditions all reach over 0.9.

| Treatment | Accuracy | Precision | Recall | F1 score |
| --- | --- | --- | --- | --- |
| Fatigue as Positive | 0.6930 | 0.8287 | 0.8089 | 0.8187 |
| Alert as Positive | 0.7573 | 0.8540 | 0.8699 | 0.8619 |
| Average | 0.7252 | 0.8414 | 0.8394 | 0.8404 |

Table 3. **Performance evaluation for framework with MSE + CE loss.** Fatigue label and alert label are regarded as positive group in turn. 69 samples are used to train the model and 91 samples are selected as test set for the calculation of accuracy, precision, recall and F1 score.

### 3.1.5 Comparisons with state-of-the-art methods

To test the superiority of our method, we compared our results with other recent approaches in facial fatigue detection using the framework with MSE + CE loss. Some of these approaches used NTHU-DDD dataset [23] and others used self-collected dataset. Most of these methods divided fatigue degree into two categories apart from the first model, which divided into four categories. The average accuracy of these methods on validation set, as shown at table 4, revealed that our method surpassed the existing 3D CNN frameworks [17, 22], and seemed to be equal to the approaches combining CNN and LSTM [21].

| Model | Implementation | Average Accuracy |
|---|---|---|
| CNN [19] | Optical flow | 73.06% |
| ResNet + LSTM [21] | SE block | 95.52% |
| CNN + LSTM [20] | Temporal smoothing | 85.52% |
| 3D-DCNN [22] | Scene understanding | 71.20% |
| 3D-CNN [17] | Gradient boosting | 87.46% |
| **3D-ResNet (ours)** | Non-local attention | **90.80%** |

Table 4. **Comparison of average accuracy on validation set under different state-of-the-art methods in detecting video-based fatigue.** Methods in [19] is a 2D model using two-stream convolutional network and convolutional network. Methods in [20, 21] are 2D models combining convolution and LSTM together. Methods in [17, 22] and ours are all 3D models. NTHU-DDD dataset is used in [17, 19, 20, 22] and self-collected dataset is used in [21]. The implementation column shows some modifications used in addition to the backbone network. The average accuracy of our model surpasses other methods except for the method in [21].

## 3.2 Sensitivity Analysis

For the framework with MSE loss, we studied its sensitivity on different training and validation sets by 5-fold cross validation. At table 5, where compared the prediction error with ground truth, the minimum error in validation set and number of epochs before early stop between fold 3 and 4 were in distinct contrast. Overall, the average minimum loss in validation set was about 0.0557.

| Cross Validation Fold | Train loss | | Validation loss | | Total Epochs |
|---|---|---|---|---|---|
| | Init | Min | Init | Min | |
| Fold 1 | 0.1095 | 0.0277 | 0.1094 | 0.0529 | 42 |
| Fold 2 | 0.1068 | 0.0242 | 0.1330 | 0.0562 | 71 |
| Fold 3 | 0.1107 | 0.0409 | 0.1003 | 0.0668 | 20 |
| Fold 4 | 0.1141 | 0.0250 | 0.1189 | 0.0483 | 80 |
| Fold 5 | 0.1063 | 0.0278 | 0.1278 | 0.0545 | 52 |

Table 5. **Change of loss values for framework with MSE loss using 5-fold cross validation.** Initial loss value and minimum loss value of training and validation set are showed for simplicity. There is a slight difference in loss values and number of epochs to trigger early stop for different folds.

Since the label of fatigue levels came from subjective judgments, there were inevitably with some errors. For the framework with MSE + CE loss, we studied its sensitivity with different samples to detect the influence of label subjectiveness. In figure 5, one was training with total 375 samples while the other was training just with 69 samples which were carefully selected with no ambiguity features between fatigue and alert state. In 375 samples, the total loss kept an increase after the first few epochs and the highest accuracy was about 75% on validation data. On the contrary, in 69 samples total loss kept decreasing after approaching 0.05 at around 30 epochs and the highest accuracy reached over 95% on validation data (Fig 4).

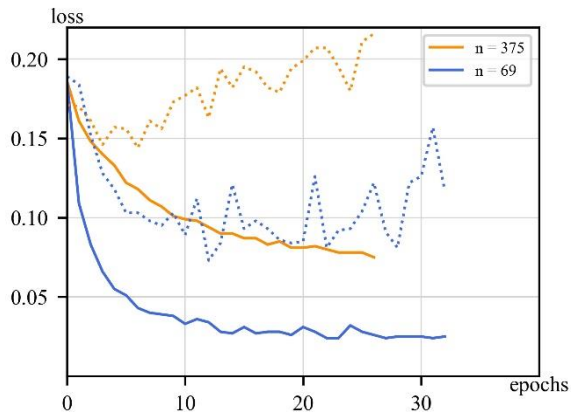

Figure 5. **Comparison of loss curve for different sample sizes during training.** That shows the loss curves using our framework with MSE + CE loss. The sample size *n* is divided into two types, namely all 375 samples and selected 69 samples. Solid and dashed lines denote loss values for training and validation set.

## 3.3 Visualization

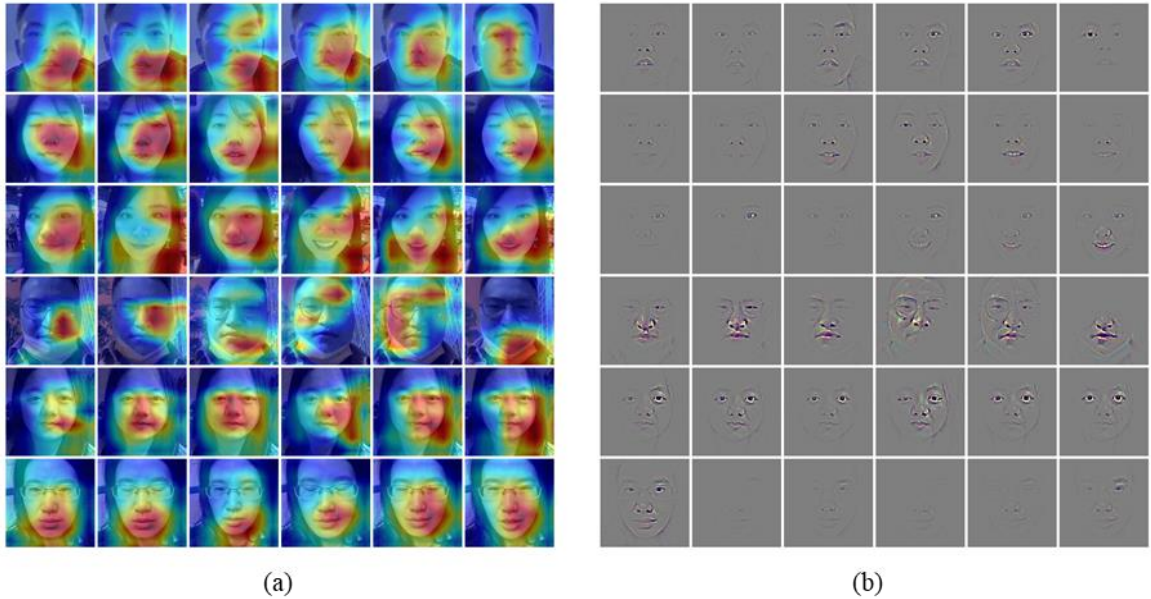

Figure 6. **Visualization of feature maps from samples in DLFD dataset using Grad-CAM and guided backpropagation**. These feature maps are extracted after the second convolution in the fourth residual block using the framework with MSE + CE loss. Different samples are shown in different rows and different frames from the same sample are shown in different columns. (a) Activation maps using Grad-CAM. That shows gradient class activation maps from 6 samples in which fatigue degree is increased from top row to bottom row. And in each sample, different clips from one video are selected to represent the features extracted by our framework. (b) Activation maps using guided backpropagation. That shows guided backpropagation maps which are corresponding to their activation maps.

In order to interpret the machine vision and analyze the concentration of neural network for detecting fatigue, we used Grad-CAM [33] to visualize the feature maps produced by our framework with MSE + CE loss. Grad-CAM has been widely used for understanding the learning process of CNN in both 2D and 3D conditions [34, 35]. The warm color (red, orange, yellow etc.) regions indicate high attention for CNN while cool color (blue, purple, gray etc.) regions indicate low attention. In our experiment, the dot product of the extracted weights from the second convolutional layer in the fourth residual block and the following feature map was calculated to produce the class activation map. Moreover, we used guided backpropagation [36] to filter out those discriminative regions on the image which positively contributed to neurons to judge fatigue. In figure 6, six images were selected from different clips in each video sample from left to right and six video samples were selected. Those images were sorted by in ascending order of fatigue degree from top to bottom.

## 4 Discussion

In our experiments, we presented a process for the optimalization of neural network parameters including batch size, data augmentation strategy, backbone model, and position of attention block. We chose batch size of 32 since experiments revealed that a large batch size significantly stabilized the learning curve and delayed the occurrence of overfitting. And the more augmentation strategy of 6 augmentations was picked due to more diversified data augmentation, which helped the network focus on the features that best responded to fatigue and exclude the influence of some noisy factors like illumination. For the choice of backbone model, 3D-ResNet indeed performed better than 3D-ResNet mainly because it could capture temporal features which image-based methods ignored. Also, non-local attention mechanism enabled the model to capture the features that are correlated in time and space. And transfer learning application in our study not only made the training process more stable, but also avoided the vanishing gradients in the deep layer on relatively small dataset.

Our results on position of attention block showed that attention block constituted an irreplaceable complementary for convolution. In traditional convolutional networks, convolution operation treats either the spatial-, temporal- or channel-wise features equally and stacks multiple layers which extract local features together to obtain more global information. From a global perspective, this approach makes the generation of global information slow and easily leads to neglect of long-range information relevance. However, self-attention module gets rid of such locality limitation and pay more attention to the information that might be correlated in a long range [26]. Our results in figure 3 clearly showed that the attention block played a role in loss value reduction during training. One of main reasons for that can be speculated from an intuitive point of view that since typical features relating fatigue such as eye blinking and yawning are continuous actions which spanned multiple frames and mobilized multiple facial muscles, the attention block can be useful for building the correlation of these muscle movements across frames. However, the relationship between the position of attention block and the degree of loss value reduction is not clear so far [26-28]. It often highly depends on other settings of hyperparameter and network structures. From a practical and convenient point of view, one non-local block inserting into one convolutional layer is sufficient for accuracy improvement of our model.

Our results on loss functions showed the effectiveness of our framework with MSE + CE loss and its outstanding performance compared with other video-based fatigue detection research. Here we only discussed video-based methods because, in contrast, though existing image-based methods which used CNN either to locate facial landmarks [13] or to extract features from eye and mouth regions [12, 14-16] have achieved up to 94% accuracy [5], these methods judged fatigue by the instantaneous state of the face presented in still images and often ignored the temporal dependencies on typical moving actions on face related to fatigue, such as eye blinking and yawning. For this reason, video-based methods with neural networks like CNN + LSTM [20, 21], two-stream models [18, 19] and 3D-CNN [17, 22] have been paid more attention. Xiao et al. [21] and Shih et al. [20] both presented similar model design to combine CNN and LSTM together, mainly due to that CNN can take responsible for facial feature extraction through single video frame and LSTM can play a role in finding temporal correlation between multiple frames. Another method suitable for video-based data is a two-stream network which regards frame images and optical flow images calculated via adjacent frames as two inputs and uses CNN to analyze both temporal and spatial features. This architecture has been already applied to detect tiredness on facial videos by Park et al. [19] and Liu et al. [18], in which the latter even achieved an outstanding accuracy about 97% on 5-class classification of fatigue states. However, those architectures were usually too complicated, which required relatively high ability to calculate and long responding time, thus that depreciated its practical value. We adopt the structure of 3D convolution because of its strong ability to extract temporal and spatial features simultaneously and easy expendability from 2D convolution as well. The effectiveness of 3D CNN has been shown by the works of Huynh et al. [17] and Yu et al. [22], the former of which achieved 87% accuracy on detecting tiredness and tireless. Differing from previous works on 3D convolution, we added non-local attention block into it, which significantly enhanced the ability of our network to continuously extract long-range specifical features. This critical adjustment made our method outperform existing 3D convolutional structures on the distinction of fatigue state.

Furthermore, we conducted sensitivity analysis on our framework using different sample sizes, training sets and validation sets. Indeed, our model was slightly sensitive to label distribution since the learning curve and model accuracy varied along with different folds as training and testing set. This difference implicated that the label distribution for divided video clips could influence on the final result. Taking into account the continuity of the fatigue state, no matter the length of the video, only one label is given to each video. However, the number of small video clips to be cut depends on the length of the large video. This is able to cause the inconsistent distribution of the training set and the testing set on the small video clips in each fold, even if stratified sampling is used to ensure the similarity of the two distributions on the large videos. Also, we found in the mild fatigue, the label had a certain degree of error due to the blurring of the boundary between fatigue and non-fatigue states. When we use continuous values to mark the fatigue level, the label becomes more subjective. Though valence and arousal are widely accepted as standard for measurement of continuous emotion degree [37, 38], the standard is unavailable in measurement of continuous fatigue degree. The lack of this measurement accounts for such label subjectiveness, amplifying the difficulty of fatigue detection in the wild, especially for mild, modern but not severe fatigue.

In addition, we explored which facial region more mattered to detect fatigue through visualization at the network. From the perspective of machine vision, the changes in activation maps and guided backpropagation maps for a specific sample can be regarded as the discriminative attention of the network [33, 36]. Some typical facial areas more relating to fatigue like the regions of eyes, mouths and facial deltoid can often be accurately identified, however, the relationship between the "flow" of the attention of the network and the fatigue levels from various samples seemed to be still lack of regularity. Despite this, we still found that our framework attempted to build a relationship between the degree of facial dynamics and fatigue levels through the analysis of the heatmaps from total samples. Specifically, the attention of network (the red region on activated maps) tended to flow to those regions where facial muscles were constantly in motion. In that way, those sober people who were characterized by rapid blinking had a higher possibility to be categorized correctly due to the rapid movement of their eyelids. In contrast, those people who were drowsy could also be classified by the feature of less eye movement, for example, the network focused on other facial moving regions like mouth regardless of eyes since the eyelids were too sticky to move in predicting the sample at the last row of figure 5.

In our work, we designed an end-to-end neural network for the prediction of fatigue state without utilizing some classic fatigue-related measurements including PERCLOS [6] and eye aspect ratio [7]. These measurements can deduce some useful indicators related to fatigue like the frequency of eye blink, which make up people's intuitive judgment on fatigue and are often called handcrafted features [39]. On the other hand, the representations extracted by deep neural networks are regarded as learning-based features. Following many research on facial expression recognition [39], the combination of handcrafted features and learning-based features might be helpful to detect many subtle and detailed features relating to mild fatigue in the wild for a further study.

## 5 Conclusion

In this study, we presented a non-local 3D attention framework for the detection of facial fatigue based on our self-collected DLFD dataset at the daily life, which gathered rich audiovisual information of fatigue state in the wild. Our proposed method could predict both the continuous and categorical degree of fatigue states. And in the binary category scenario it was indicated to outperform various existing frameworks for detecting video-based fatigue. In the future work, we would like to further modify our model in terms of its accuracy with a larger dataset and also its sensitivity to mild fatigue. In addition, we are attempting the deployment of proposed algorithm on mobile phones. We expect our DLFD dataset can be further analyzed and contribute to mild fatigue detection in daily.

## Acknowledgments

This work was supported by the Key-Area Research and Development Program of Guangdong Province (2018B030331001), International Postdoctoral Exchange Fellowship Program by the Office of China Postdoctoral Council (20160021), International Partnership Program of Chinese Academy of Sciences (172644KYS820170004), Commission on Innovation and Technology in Shenzhen Municipality of China (JCYJ20150630114942262), Hong Kong, Macao, and Taiwan Science and Technology Cooperation Innovation Platform in Universities in Guangdong Province (2013gjhz0002), and the National Key R&D Program of China (2017YFC1310503). In particular, we would like to thank the Venture Mentoring Service (VMS) at MIT, Cambridge, MA, USA.

Disclosure: Z. Chen, None; X. Zhang, None; J. Li, None; G. Chen, None; S. Wang, None; F. Fan, None; C. C. Wang, None; X. LI None.